\documentclass[10pt,twocolumn,letterpaper]{article}

\usepackage{arxiv_iccv}
\usepackage{times}
\usepackage{epsfig}
\usepackage{graphicx}
\usepackage{amsmath}
\usepackage{amssymb}
\usepackage{bm}
\usepackage{mathrsfs}
\usepackage{diagbox}
\usepackage{makecell}
\usepackage{xcolor}

\newcommand{\SemNei}{\mathbf{C}_i}
\newcommand{\ConNei}{\mathbf{B}_i}
\newcommand{\VI}{\mathbf{v}_i}
\newcommand{\BmTheta}{\bm{\theta}}
\newcommand{\XI}{\mathbf{x}_i}
\newcommand{\LevAgr}{L(\SemNei, \ConNei | \BmTheta, \XI)}
\newcommand{\MemBank}{\mathbf{\bar{V}}}
\newcommand{\MemBankVec}{\mathbf{\bar{v}}}
\newcommand{\WEB}{\texttt{[WEBSITE\_WITHHOLD]}}
\newcommand{\Fst}{$_\mathrm{Fast}$}
\newcommand{\Fstr}{$_\mathrm{Faster}$}

\usepackage[pagebackref=true,breaklinks=true,letterpaper=true,colorlinks,bookmarks=false,hypertexnames=false]{hyperref}

\iccvfinalcopy

\def\iccvPaperID{4162}

\ificcvfinal\fi
\begin{document}

\title{Local Aggregation for Unsupervised Learning of Visual Embeddings}

\author{Chengxu Zhuang\\
Stanford University
\and
Alex Lin Zhai\\
Stanford University
\and
Daniel Yamins\\
Stanford University
}
\date{}
\maketitle

\begin{abstract}
Unsupervised approaches to learning in neural networks are of substantial interest for furthering artificial intelligence, both because they would enable the training of networks without the need for large numbers of expensive annotations, and because they would be better models of the kind of general-purpose learning deployed by humans.  However, unsupervised networks have long lagged behind the performance of their supervised counterparts, especially in the domain of large-scale visual recognition.  Recent developments in training deep convolutional embeddings to maximize non-parametric instance separation and clustering objectives have shown promise in closing this gap. Here, we describe a method that trains an embedding function to maximize a metric of local aggregation, causing similar data instances to move together in the embedding space, while allowing dissimilar instances to separate.  This aggregation metric is dynamic, allowing soft clusters of different scales to emerge.  We evaluate our procedure on several large-scale visual recognition datasets, achieving state-of-the-art unsupervised transfer learning performance on object recognition in ImageNet, scene recognition in Places 205, and object detection in PASCAL VOC. 
\end{abstract}

\section{Introduction} \label{set:intro}
Deep convolutional neural networks (DCNNs) have achieved great success on many tasks across a variety of domains, such as vision~\cite{krizhevsky2012imagenet,simonyan2014very,he2016deep,he2017mask,carreira2017quo}, audition~\cite{hinton2012deep,hannun2014deep,deng2013new,noda2015audio}, and natural language processing~\cite{young2018recent,hirschberg2015advances,conneau2016very,kumar2016ask}.
However, most successful DCNNs are trained in a supervised fashion on labelled datasets~\cite{krizhevsky2012imagenet,simonyan2014very,he2016deep,deng2009imagenet,hinton2012deep}, requiring the costly collection of large numbers of annotations.
There is thus substantial interest in finding methods that can train DCNNs solely using unlabeled data, which are often readily available. 
Over many decades of work, substantial progress has been achieved using a wide variety of unsupervised learning approaches~\cite{caron2018deep, wu2018unsupervised, zhang2017split,krahenbuhl2015data,doersch2015unsupervised, donahue2016adversarial,zhang2016colorful,noroozi2016unsupervised,wang2015unsupervised,noroozi2017representation}. Nevertheless, unsupervised networks are still typically significantly lower performing than their supervised counterparts, and are rarely used in real-world applications~\cite{caron2018deep,noda2015audio,carreira2017quo}.

In contrast to the inefficiency of unsupervised learning in artificial neural networks, humans and non-human primates develop powerful and domain-general visual systems with very few labels~\cite{lewis2005multiple,braddick2011development,wattam2010reorganization,atkinson2002developing,harwerth1986multiple,bourne2005hierarchical,sporns2004organization}. 
Although the mechanisms underlying the efficiency of biological learning still remain largely unknown~\cite{braddick2011development}, researchers reliably report that infants as young as three months can group perceptually similar stimuli~\cite{mareschal2001categorization}, even for stimulus types that the infants have never seen before.
Moreover, this ability arises long before these infants appear to have an explicit concept of object category~\cite{mareschal2001categorization,rakison2003early,husaim1981infant,cohen1979concept}. 
These findings suggest that biological unsupervised learning may take advantage of inherent visual similarity, without requiring sharp boundaries between stimulus categories.

Inspired by these results, we propose a novel unsupervised learning algorithm through local non-parametric aggregation in a latent feature space.
First, we non-linearly embed inputs in a lower-dimensional space via a neural network.
We then iteratively identify close neighbors surrounding each example in the embedding space, while optimizing the embedding function to strengthen the degree of local aggregation. 
Our procedure, which we term Local Aggregation (LA), causes inputs that are naturally dissimilar to each other to move apart in the embedding space, while allowing inputs that share statistical similarities to arrange themselves into emergent clusters.
By simultaneously optimizing this soft clustering structure and the non-linear embedding in which it is performed, our procedure exposes subtle statistical regularities in the data.
The resulting representation in turn robustly supports downstream tasks.

Here, we illustrate the LA procedure in the context of large-scale visual learning.
Training a standard convolution neural network with LA using images from ImageNet~\cite{deng2009imagenet} significantly outperforms current state-of-art unsupervised algorithms on transfer learning to classification tasks on both ImageNet and the Places 205 dataset~\cite{zhou2014learning}.
In addition, LA shows consistent improvements as the depth of the embedding function increases, allowing it to achieve $\mathbf{60.2\%}$ top-1 accuracy on ImageNet classification. This is, as far as we know, the first time an unsupervised model has surpassed the milestone AlexNet network trained directly on the supervised task.
We also show that, through further fine-tuning, LA trained models obtain state-of-the-art results on the PASCAL object detection task. 

The remainder of this paper is organized as follows: in section~\ref{set:relat}, we discuss related work; in section~\ref{set:method}, we describe the LA method; in section~\ref{sec:results}, we show experimental results; in section~\ref{sec:analysis}, we present analyses illustrating how this algorithm learns and justifying key parameter choices.



\section{Related Work} \label{set:relat}
Unsupervised learning methods span a very broad spectrum of approaches going back to the roots of artificial neural networks~\cite{radford2015unsupervised,le2011building,barlow1989unsupervised,sanger1989optimal,hinton2006fast, lee2009convolutional,hinton2006reducing,hopfield1982neural,hebb1961organization}, and are too numerous to fully review here. 
However, several recent works have achieved exciting progress in unsupervised representation learning~\cite{caron2018deep, wu2018unsupervised, zhang2017split}.
Although the LA method draws inspiration from these works, it differs from them in some important conceptual ways.

\textbf{DeepCluster.}
DeepCluster~\cite{caron2018deep} (DC) trains a DCNN in a series of iterative rounds.  In each round, features from the penultimate layer of the DCNN from the previous round are clustered, and the cluster assignments are used as self-generated supervision labels for further training the DCNN using standard error backprogation. 
Like DC, LA also uses an iterative training procedure, but the specific process within each iteration differs significantly.
First, unlike the clustering step of DC where all examples are divided into mutually-exclusive clusters, our method identifies neighbors separately for each example, allowing for more flexible statistical structures than a partition. 
Indeed, as shown in Section~\ref{sec:hyper}, the use of individual semantic neighbor identifiers rather than global clustering is important for performance improvement.
Secondly, the optimization step of LA differs from that of DC by optimizing a different objective function.
Specifically, DC optimizes the cross-entropy loss between predicted and ground truth cluster labels, requiring an additional and computationally expensive linear readout layer.
Moreover, due to arbitrary changes in the cluster label indices across iterative rounds, this additional readout layer needs to be frequently recomputed.
In contrast, LA employs an objective function that directly optimizes a local soft-clustering metric, requiring no extra readout layer and only a small amount of additional computation on top of the feature representation training itself.  
These differences lead both to better final performance and substantially improved training efficiency. 

\textbf{Instance Recognition.}
The Instance Recognition task~\cite{wu2018unsupervised} (IR) treats each example as its own ``category'' and optimizes the DCNN representation to output an embedding in which all examples are well-separated from each other. 
LA uses a similar embedding framework, but achieves significantly better performance by pursuing a distinct optimization goal. 
Specifically, while IR optimizes for equally separating representations of all examples, LA encourages a balance between separation and clustering on a per-example basis, as measured by the local aggregation criterion. 
For this reason, the LA approach can be thought of as a principled hybrid between the DC and IR approaches.

\textbf{Self-supervised ``missing-data'' tasks.}
These tasks build representations by hiding some information about each example input, and then optimizing the network to predict the hidden information from the visible information that remains. 
Examples include context prediction~\cite{doersch2015unsupervised}, colorization of grayscale images~\cite{doersch2015unsupervised}, inpainting of missing portions of images~\cite{pathak2016context}, and the Split-Brain method~\cite{zhang2017split}.
However, it is ultimately unclear whether these tasks are perfectly aligned with the needs of robust visual representation.
Indeed, it has been found that deeper networks better minimizing the loss functions used in such tasks gain little transfer learning performance on object recognition tasks~\cite{doersch2017multi}. 
Moreover, most missing-data tasks rely on structures that are specific to visual data, making them potentially less general than the embedding/clustering concepts used in DC, IR or our LA method.

\textbf{Generative models}.
Another broad class of unsupervised learning algorithm, often termed deep generative models, focuses on reconstructing input images from a bottlenecked latent representation.
The networks trained by these algorithms use the latent representations for other tasks, including object recognition. 
These learning methods include classical ones such as Restricted Boltzman Machines~\cite{hinton2006fast, lee2009convolutional} as well as more recent ones such as Variational Auto-Encoders~\cite{kingma2013auto} and Generative Adversarial Networks~\cite{donahue2016adversarial, goodfellow2014generative}.
Although the features learned by generative models have been put to a wide variety of exciting uses~\cite{ledig2017photo,zhang2017stackgan,denton2015deep,li2016precomputed,jetchev2016texture}, their power as latent representations for downstream visual tasks such as object recognition has yet to be fully realized.

\section{Methods} \label{set:method}
\begin{figure*}
\begin{center}
\includegraphics[width=\textwidth] {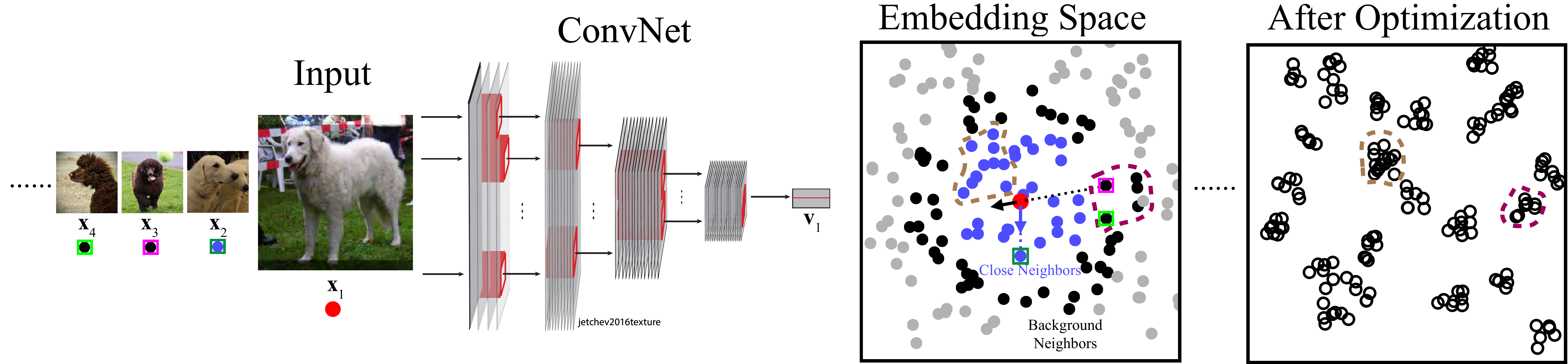}
\end{center}
\caption{
Illustration of the Local Aggregation (LA) method.
For each input image, we use a deep neural network to embed it into a lower dimension space ("Embedding Space" panel).
We then identify its close neighbors (blue dots) and background neighbors (black dots).
The optimization seeks to push the current embedding vector (red dot) closer to its close neighbors and further from its background neighbors.
The blue arrow and black arrow are examples of influences from different neighbors on the current embedding during optimization.
The "After Optimization" panel illustrates the typical structure of the final embedding after training.
}
\label{fig:schema}
\end{figure*}

Our overall objective is to learn an embedding function $f_{\BmTheta}$ (realized via a neural network) that maps images $\mathbf{I} = \{\mathbf{x}_1, \mathbf{x}_2, ..., \mathbf{x}_N\}$ to features $\mathbf{V} = \{ \mathbf{v}_1, \mathbf{v}_2, ..., \mathbf{v}_N\}$ with $\VI = f_{\BmTheta} (\XI)$ in a compact $D$-dimension representation space where similar images are clustered while dissimilar images are separated.
To achieve this objective, we design an iterative procedure to bootstrap the aggregation power of a deep non-linear embedding function. 
More specifically, at any given stage during training the embedding function, we dynamically identify two sets of neighbors for an $\XI$ and its embedding $\VI$: \emph{close neighbors} $\SemNei$ and \emph{background neighbors} $\ConNei$.  Intuitively, close neighbors are those whose embeddings should be made similar to $\VI$, while background neighbors are used to set the distance scale with respect to which the judgement of closeness should be measured. To help better understand these two sets, we provide a schematic illustration in Fig.~\ref{fig:schema}, and describe the details of how they are defined mathematically in section~\ref{sec:nei_iden}.  Using $\ConNei$ and $\SemNei$, we then define the level of \textit{local aggregation} $\LevAgr$ near each input $\XI$, which characterizes the relative level of closeness within $\SemNei$, compared to that in $\ConNei$.
The parameters $\BmTheta$ of the neural network realizing the embedding function are then tuned over the course of training to maximize $\LevAgr$. 

\subsection{Neighbor Identification}\label{sec:nei_iden}
We first describe how the neighbor types $\ConNei$ and $\SemNei$ are defined. \textit{Nearest-neighbor based identification for $\ConNei$}:  
At any given step of optimization, the background neighbors for a given embedded point $\VI$  are simply defined as the $k$ closest embedded points $\mathcal{N}_k(\VI)$ within $\mathbf{V}$, where distance is judged using the cosine distance on the embedding space. 
The number $k$ of background neighbors to be used is a hyperparameter of the algorithm.  \textit{Robustified clustering-based identification for $\SemNei$}:
To identify close neighbors, we first apply an unsupervised clustering algorithm on all embedded points $\mathbf{V}$ to cluster the representations into $m$ groups $\mathbf{G} = \{ G_1, G_2, ..., G_m\}$.
Let $g(\VI)$ denote the cluster label of $\VI$ in this clustering result, i.e. $i \in G_{g(\VI)}$.  
In the simplest version of our procedure, we then define $\SemNei$ to be the set $G_{g(\VI)}$. However, because clustering can be a noisy and somewhat arbitrary process, we compute multiple clusterings under slightly different conditions, and then aggregate neighbors across these multiple clusterings to achieve more stable results.  
Specifically, let $\{ \mathbf{G}^{(j)} \}$ be clusters for $H$ distinct clusterings, where $\mathbf{G}^{(j)} = \{ G_1^{(j)}, G_2^{(j)}, ..., G_{m^{(j)}}^{(j)}\}$ with $j \in \{1,2,...,H\}$, and $\{g^{(j)}\}$ defined accordingly. 
We then define $\SemNei = \bigcup_{j=1}^{H} G_{g^{(j)}(\VI)}^{(j)}$.
The number $m$ of clusters and number $H$ of clusterings are hyperparameters of the algorithm. In this work, we use $k$-means clustering as the standard unsupervised algorithm.

Intuitively, background neighbors are an unbiased sample of nearby points that (dynamically) set the scale at which ``close-ness'' should be judged; while close neighbors are those that are especially nearby, relative to those in other clusters. 
The mathematical definitions above represent just one specific way to formalize these ideas, and many alternatives are possible.
In Section~\ref{sec:hyper}, we show that our choices are not arbitrary by exploring the consequences of making alternate decisions.  

\subsection{Local Aggregation Metric}
Given the definition of $\ConNei$ and $\SemNei$, we describe the formulation of our local aggregation metric, $\LevAgr$. We build our formulation upon a non-parametric softmax operation proposed by Wu et al. in~\cite{wu2018unsupervised}.
In that work, the authors define the probability that an arbitrary feature $\mathbf{v}$ is recognized as the $i$-th image to be: 
\begin{equation}
    P(i | \mathbf{v}) = \frac{\mathrm{exp}(\VI^T \mathbf{v} / \tau)}{\sum_{j=1}^N\mathrm{exp}(\mathbf{v}_j^T \mathbf{v} / \tau)}
    \label{equ:single}
\end{equation}
where $\tau \in [0, 1]$ is a fixed scale hyperparameter, and where both $\{\VI\}$ and $\mathbf{v}$ are projected onto the L2-unit sphere in the $D$-dimensional embedding space (e.g. normalized such that $\| \mathbf{v} \|_2 = 1$).

Following equation~\ref{equ:single}, given an image set $\mathbf{A}$, we then define the probability of feature $\mathbf{v}$ being recognized as an image in $\mathbf{A}$ as:
\begin{equation}
    P(\mathbf{A} | \mathbf{v}) = \sum_{i \in \mathbf{A}} P(i | \mathbf{v})
    \label{equ:set}
\end{equation}

Finally, we formulate $\LevAgr$ as the negative log-likelihood of $\VI$ being recognized as a close neighbor (e.g. is in $\SemNei$), given that $\VI$ is recognized as a background neighbor (e.g. is in $\ConNei$):
\begin{equation}
    \LevAgr = - \mathrm{log} \frac {P(\SemNei \cap \ConNei | \VI)} {P(\ConNei | \VI)}
    \label{equ:l_eq}
\end{equation}

The loss to be minimized is then:
\begin{equation}
    \mathcal{L}_i = \LevAgr + \lambda \| \BmTheta \|^2_2
    \label{equ:loss}
\end{equation}
where $\lambda$ is a regularization hyperparameter. 

\textbf{Discussion}. Because the definition of $\LevAgr$ is somewhat involved, we describe a simple conceptual analysis that illustrates the intuition for why we chose it as a measure of local aggregation. 
Letting $\SemNei^c$ denote the complement of $\SemNei$ in $\mathbf{I}$,  
we have $P(\ConNei | \VI) = P(\SemNei^c \cap \ConNei | \VI) + P(\SemNei \cap \ConNei | \VI)$. Thus, from equation~\ref{equ:l_eq}, we see that $\LevAgr$ is minimized when $P(\SemNei \cap \ConNei | \VI)$ is maximized and $P(\SemNei^c \cap \ConNei | \VI)$ is minimized.  
It is easy to understand the meaning of minimizing $P(\SemNei^c \cap \ConNei | \VI)$: this occurs as the distances between $\VI$ and its non-close background neighbors are maximized.  
The consequences of maximizing $P(\SemNei \cap \ConNei | \VI)$ are a bit more subtle. 
As shown empirically in~\cite{wu2018improving} (albeit in the supervised context), as long as the scaling parameter $\tau \ll 1$, maximizing $P(\mathbf{A} | \VI)$ for any set $\mathbf{A}$ causes the emergence of natural ``sub-categories'' in (the embeddings of) $\mathbf{A}$, and encourages $\VI$ to move closer to \emph{one} of these sub-categories rather than their overall average.
This empirical result can be intuitively understood by recognizing the fact that $\mathrm{exp}(\VI^T \mathbf{v} / \tau)$ increases exponentially when $\VI^T \mathbf{v}$ approaches 1, suggesting that $P(\mathbf{A} | \VI)$ will approach 1 when $\mathbf{A}$ includes a small cluster of features that are all very close to $\mathbf{v}$.
Putting these observations together, the optimized representation space created by minimizing $\LevAgr$ is, intuitively, like that shown in Fig.~\ref{fig:schema}: a set of embedded points that have formed into small clusters at a distribution of natural scales. 

\subsection{Memory Bank}\label{sec:mem_bank}
As defined above, the neighbor identification procedures and the loss function implicitly describe computations involving all the embedded features $\mathbf{V}$, which soon becomes intractable for large datasets.
To address this issue, we follow~\cite{wu2018unsupervised, wu2018improving} and maintain a running average for $\mathbf{V}$, which is called the \emph{memory bank}, denoted $\MemBank = \{ \MemBankVec_1, \MemBankVec_2, ..., \MemBankVec_N\}$.
Similarly to \cite{wu2018unsupervised, wu2018improving}, we initialize the memory bank with random $D$-dimensional unit vectors and then update its values by mixing $\MemBankVec_i$ and $\mathbf{v}_i$ during training as follows:
\begin{equation}
    \MemBankVec_i \gets (1-t) \MemBankVec_i + t \mathbf{v}_i
    \label{equ:memBank}
\end{equation}
where $t$ $\in [0, 1]$ is a fixed mixing hyperparameter. 
With the help of $\MemBank$, we can then rewrite the neighbor identification procedures and equation~\ref{equ:single} by replacing the feature sets $\mathbf{V}$ with $\MemBank$.
In particular for $\SemNei$, the cluster label function $g$ is applied to $\MemBankVec_i$ by index identification, ensuring the chosen cluster includes the index $i$ itself.
After this replacement, it is no longer necessary to recompute $\mathbf{V}$ before every step to identify (good approximations of) $\SemNei$ and $\ConNei$.

\section{Results} \label{sec:results}
In this section, we describe tests of the LA method on visual representation learning and compare its performance to that of other methods.

\subsection{Experiment Settings}\label{sec:expset}
We first list key parameters used for network training.
Following~\cite{wu2018unsupervised}, we set parameter $\tau = 0.07$, $D = 128$, $\lambda = 0.0001$, and $t = 0.5$. 
For all network structures, we use SGD with momentum of 0.9 and batch size 128. 
Initial learning rates are set to 0.03, and dropped by a factor of 10 when validation performances saturate, typically leading to training for 200 epochs with two learning rate drops.
Most of these parameters are taken from~\cite{wu2018unsupervised}, as our conceptual framework is similar, but a further hyper-parameter search might lead to better results, given that our optimization goal differs substantially.

As a warm start for our models, we begin training using the IR loss function for the first 10 epochs, before switching over to using the LA method.
Following the methods of \cite{caron2018deep}, for AlexNet~\cite{krizhevsky2012imagenet} and VGG16~\cite{simonyan2014very} architectures, 
we add batch normalization (BN) layers~\cite{ioffe2015batch} after all convolution and fully-connected layers, before \texttt{ReLu} operations, to allow a higher learning rate and a faster convergence speed.
Though adding BN is known to improve convergence speed but not typically to lead to higher final ImageNet performance levels using supervised training regimes, it is unclear whether this remains true when using unsupervised training methods.
Importantly, the potentially competitive IR method~\cite{wu2018unsupervised} did not originally include BN in their AlexNet and VGG16, so to ensure that we have fairly compared that method to LA or DC, we also train AlexNet and VGG16 with BN on the IR task.
For all structures, we replace the final category readout layer with a linear layer with $D$ output units, followed by a L2-normalization operation to ensure that the output is a unit vector.

We set $k=4096$ for computing $\ConNei$ using the nearest neighbors procedure. 
In computing $\SemNei$, we use $k$-means~\cite{lloyd1982least} implemented in Faiss~\cite{JDH17} as the standard unsupervised clustering algorithm, generating multiple clusterings for robustness via different random initializations.
Using the notation of Section \ref{set:method}, AlexNet is trained with $H=3, m=30000$, VGG16 is trained with $H=6, m=10000$, all ResNet structures are trained with $H=10, m=30000$.
We justify all parameter choices and intuitively explain why they are optimal in Section~\ref{sec:hyper}.
All code for reproducing our training is available at: \WEB.

\subsection{Transfer Learning Results} \label{sec:trans}
After fully training networks on ImageNet, we then test the quality of the learned visual representations by evaluating transfer learning to other tasks, including ImageNet classification on held-out validation images, scene classification on Places205~\cite{zhou2014learning}, and object detection on PASCAL VOC 2007~\cite{everingham2010pascal}.
For classification tasks, we also report K-nearest neighbor (KNN) classification results using the embedding features, acquired via a method similar to that in~\cite{wu2018unsupervised}.
Specifically, we take top $K$ nearest neighbors $\mathcal{N}_K$ for the feature $\mathbf{v}$ either (for ImageNet) from the saved memory bank or (for Places) from the computed network outputs for center crops of training images.
Their labels are then weighted by $\mathrm{exp}(\VI^T \mathbf{v} / \tau)$ and combined to get final predictions. We report results for $K = 200$ as in~\cite{wu2018unsupervised}.

\textbf{Object Recognition.}
To evaluate transfer learning for the ImageNet classification task, we fix network weights learned during the unsupervised procedure, add a linear readout layer on top of each layer we want to evaluate, and train the readout using cross-entropy loss together with L2 weight decay.
We use SGD with momentum of 0.9, batch size 128, and weight decay $0.0001$. 
Learning rate is initialized at 0.01 and dropped by a factor of 10 when performance saturates, typically leading to 90 training epochs with two learning rate drops.
We report 10-crop validation performances to ensure comparability with ~\cite{caron2018deep}.
Performance results in Table~\ref{tab:imagenet} show that LA significantly outperforms other methods with all architectures, especially in deeper architectures.
LA-trained AlexNet reaches $42.4\%$, which is $1.4\%$ higher than previous state-of-the-art.
Improvements over previous unsupervised state-of-the-art are substantially larger for VGG16 ($+4.9\%$), ResNet-18 ($+3.7\%$), and ResNet-50 ($+6.2\%$).
In particular, LA-trained ResNet-50 achieves $\mathbf{60.2\%}$ top-1 accuracy on ImageNet classification, surpassing AlexNet trained directly on the supervised task.
Using KNN classifiers, LA outperforms the IR task by a large margin with all architectures.
There is a consistent performance increase for the LA method both from overall deeper architectures, and from earlier layers to deeper layers within an architecture. 
Most alternative training methods (e.g. \cite{pathak2017learning, malisiewicz2011ensemble, doersch2015unsupervised, zhang2016colorful}) do not benefit significantly from increasing depth.
For example, ResNet-101 trained using Color~\cite{zhang2016colorful} can only achieve $39.6\%$ and the best performance using ResNet-101 with unsupervised task is only $48.7\%$ with CPC~\cite{oord2018representation}.

\setlength{\tabcolsep}{3pt}
\setlength{\fboxrule}{1pt}%
\begin{table}
\begin{center}
\begin{tabular}{c|ccccc|c}
\hline
Method & conv1 & conv2 & conv3 & conv4 & conv5 & KNN \\
\hline\hline
\multicolumn{7}{c}{AlexNet}\\
\hline
Random & 11.6 & 17.1 & 16.9 & 16.3 & 14.1 & 3.5 \\
Context~\cite{doersch2015unsupervised} & 16.2 & 23.3 & 30.2 & 31.7 & 29.6 & -- \\
Color~\cite{zhang2016colorful} & 13.1 & 24.8 & 31.0 & 32.6 & 31.8 & -- \\
Jigsaw~\cite{noroozi2016unsupervised} & \textbf{19.2} & 30.1 & 34.7 & 33.9 & 28.3 & -- \\
Count~\cite{noroozi2017representation} & 18.0 & 30.6 & 34.3 & 32.5 & 25.7 & -- \\
SplitBrain~\cite{zhang2017split} & 17.7 & 29.3 & 35.4 & 35.2 & 32.8 & 11.8 \\
IR~\cite{wu2018unsupervised} & 16.8 & 26.5 & 31.8 & 34.1 & 35.6 & 31.3 \\
IR(with BN)* & 18.4 & 30.1 & 34.4 & 39.2 & 39.9 & 34.9 \\
DC~\cite{caron2018deep} & 13.4 & 32.3 & \textbf{41.0} & 39.6 & 38.2 & -- \\
LA (ours) & 18.7 & \textbf{32.7} & 38.1 & \textbf{42.3} & \fboxsep=1pt\fcolorbox{red}{white}{\textbf{42.4}} & \textbf{38.1} \\
\hline
\multicolumn{7}{c}{VGG16}\\
\hline
IR & 16.5 & 21.4 & 27.6 & 35.1 & 39.2 & 33.9 \\
IR(with BN)* & 13.2 & 18.7 & 27.3 & 39.8 & 50.4 & 42.1 \\
DC* & \textbf{18.2} & \textbf{27.5} & \textbf{41.5} & \textbf{51.3} & 52.7 & -- \\
LA (ours) & 14.3 & 23.4 & 28.3 & 44.5 & \fboxsep=1pt\fcolorbox{red}{white}{\textbf{57.6}} & \textbf{46.6} \\
\hline
\multicolumn{7}{c}{ResNet-18}\\
\hline
IR & 16.0 & \textbf{19.9} & 29.8 & 39.0 & 44.5 & 41.0 \\
DC* & \textbf{16.4} & 17.2 & 28.7 & 44.3 & 49.1 & -- \\
LA (ours) & 9.1 & 18.7 & \textbf{34.8} & \textbf{48.4} & \fboxsep=1pt\fcolorbox{red}{white}{\textbf{52.8}} & \textbf{45.0} \\
\hline
\multicolumn{7}{c}{ResNet-50}\\
\hline
IR & 15.3 & 18.8 & 24.9 & 40.6 & 54.0 & 46.5 \\
DC* & \textbf{18.9} & \textbf{27.3} & 36.7 & \textbf{52.4} & 44.2 & -- \\
LA (ours) & 10.2 & 23.3 & \textbf{39.3} & 49.0 & \fboxsep=1pt\fcolorbox{red}{white}{\textbf{60.2}} & \textbf{49.4} \\
\hline
\end{tabular}
\end{center}
\caption{
ImageNet transfer learning and KNN classifier performance. 
Numbers within the red box are the best for the given architecture.
Performances of most methods using AlexNet are taken from~\cite{caron2018deep, wu2018unsupervised}.
*: performance number produced by us, please refer to the supplementary material for training details.
}
\label{tab:imagenet}
\end{table}

\textbf{Scene Categorization.}
To test the generalization ability of the learned representations to a data distribution distinct from that used in training, we assessed transfer to the Places~\cite{zhou2014learning} dataset, which includes 2.45$M$ images labelled with 205 scene categories.
As in the previous section, we train linear readout layers for the scene categorization task on top of the pretrained ImageNet model, using training procedures and hyper-parameters identical to those used in ImageNet transfer learning.
Results shown in Table~\ref{tab:places} illustrate that the LA method surpasses previous methods in transfer learning performance with all architectures, especially with deeper networks. 
Please refer to the supplementary material for $K$-nearest neighbor classification performance.
These result indicate strong generalization ability of the visual representations learned via the LA method.

\setlength{\tabcolsep}{3pt}
\begin{table}
\begin{center}
\begin{tabular}{c|ccccc}
\hline
Method & conv1 & conv2 & conv3 & conv4 & conv5 \\
\hline\hline
\multicolumn{6}{c}{AlexNet}\\
\hline
Random & 15.7 & 20.3 & 19.8 & 19.1 & 17.5 \\
Context~\cite{doersch2015unsupervised} & 19.7 & 26.7 & 31.9 & 32.7 & 30.9 \\
Color~\cite{zhang2016colorful} & 22.0 & 28.7 & 31.8 & 31.3 & 29.7 \\
Jigsaw~\cite{noroozi2016unsupervised} & \textbf{23.0} & 32.1 & 35.5 & 34.8 & 31.3 \\
SplitBrain~\cite{zhang2017split} & 21.3 & 30.7 & 34.0 & 34.1 & 32.5 \\
IR~\cite{wu2018unsupervised} & 18.8 & 24.3 & 31.9 & 34.5 & 33.6 \\
IR(with BN)* & 21.3 & 33.0 & 36.5 & 39.2 & 38.7 \\
DC~\cite{caron2018deep} & 19.6 & \textbf{33.2} & \textbf{39.2} & 39.8 & 34.7 \\
LA (ours) & 18.7 & 32.7 & 38.2 & \fboxsep=1pt\fcolorbox{red}{white}{\textbf{40.3}} & \textbf{39.5} \\
\hline
\multicolumn{6}{c}{VGG16}\\
\hline
IR & 17.6 & 23.1 & 29.5 & 33.8 & 36.3 \\
IR(with BN)* & 17.3 & 22.9 & 27.3 & 39.3 & 45.8 \\
DC* & \textbf{21.5} & \textbf{31.6} & \textbf{40.9} & \textbf{45.2}  & 44.2  \\
LA (ours) & 20.1 & 25.9 & 31.9 & 44.0 & \fboxsep=1pt\fcolorbox{red}{white}{\textbf{50.0}} \\
\hline
\multicolumn{6}{c}{ResNet-18}\\
\hline
IR & 17.8 & 23.0 & 30.1 & 37.0 & 38.1 \\
DC* & 16.4 & 22.5 & 30.5 & 40.4 & 41.8 \\
LA (ours) & \textbf{18.9} & \textbf{26.7} & \textbf{36.5} & \textbf{44.7} & \fboxsep=1pt\fcolorbox{red}{white}{\textbf{45.6}} \\
\hline
\multicolumn{6}{c}{ResNet-50}\\
\hline
IR & 18.1 & 22.3 & 29.7 & 42.1 & 45.5 \\
DC* & \textbf{20.1} & \textbf{29.1}  & 35.3 & 43.2 & 38.9 \\
LA (ours) & 10.3 & 26.4 & \textbf{39.9} & \textbf{47.2} & \fboxsep=1pt\fcolorbox{red}{white}{\textbf{50.1}} \\
\hline
\end{tabular}
\end{center}
\caption{
Places transfer learning performance.
*: performances produced by us, please refer to the supplement for details. 
}
\label{tab:places}
\end{table}

\textbf{Object Detection.}
The results presented in Table~\ref{tab:imagenet} and~\ref{tab:places} illustrate the utility of LA for learning representations for visual categorization tasks.
However, visual challenges faced in real life also include other tasks, such as object detection.
Therefore, we also evaluate the transfer learning ability of our models to the object detection task in the PASCAL VOC 2007~\cite{everingham2010pascal} dataset.
The typical PASCAL detection task evaluation procedure~\cite{caron2018deep, zhang2017split, wu2018unsupervised, wang2017transitive} fine-tunes unsupervised architectures using the Fast RCNN~\cite{girshick2015fast} method. 
However, Fast RCNN is substantially less computationally efficient than more recently proposed pipelines such as Faster RCNN~\cite{renNIPS15fasterrcnn} or Mask RCNN~\cite{he2017mask}, and is less well-supported by validated reference implementations in common deep learning frameworks.
To ensure training efficiency and correctness, in this work we have used the Faster RCNN pipeline from validated implementations in both TensorFlow and Pytorch.
However, because the performance achieved by Faster RCNN can vary somewhat from that of Fast RCNN, direct comparison of these results to numbers generated with Fast RCNN may be misleading.
For this reason, we have additionally evaluated models trained with IR and DC using Faster RCNN where possible.
For implementation details, please refer to the supplementary material.
Results are shown in Table~\ref{tab:pascal}, illustrating that the LA method achieves state-of-the-art unsupervised transfer learning for the PASCAL detection task. Interestingly, the performance gaps between the best unsupervised methods and the supervised controls are comparatively smaller for the PASCAL task than for the classification tasks.

\setlength{\tabcolsep}{3pt}
\begin{table}
\begin{center}
\begin{tabular}{c|cc|cc|c}
\hline
Method & A\Fst & A\Fstr & V\Fst & V\Fstr & R\Fstr \\
\hline
\hline
Supervised & 56.8 & 54.3 & 67.3 & 70.0 & 74.6 \\
\hline
Jigsaw~\cite{noroozi2016unsupervised} & 53.2 & -- & --  & -- & -- \\
Video~\cite{wang2015unsupervised} & 47.2 & -- & 60.2 & -- & -- \\
Context~\cite{doersch2015unsupervised} & 51.1 & -- & 61.5 & -- & -- \\
Trans~\cite{wang2017transitive} & -- & -- & 63.2 & -- & -- \\
IR~\cite{wu2018unsupervised} & 48.1 & 53.1 & 60.5 & 65.6 & 65.4 \\
DC & \textbf{55.4} & -- & \textbf{65.9} & -- & -- \\
\hline
LA (ours) & -- & \textbf{53.5} & -- & \textbf{68.4} & \textbf{69.1} \\
\hline
\end{tabular}
\end{center}
\caption{
PASCAL VOC 2007 detection mAP. 
A=AlexNet, V=VGG16, and R=ResNet50.
Bold numbers are the best in their columns.
Performances with Faster RCNN are produced by us, except that of ResNet50 of IR, which is as reported in~\cite{wu2018unsupervised}.
Most numbers using Fast RCNN are taken from~\cite{caron2018deep, wu2018unsupervised}.
For numbers produced by us, we show the averages of three independent runs.  Standard deviations are close to $0.2\%$ in all cases.
}
\label{tab:pascal}
\end{table}

\section{Analysis} \label{sec:analysis}
\subsection{Visualizations}
In this subsection, we analyze the embedding space through visualizations.

\begin{figure}[t]
\begin{center}
\includegraphics[width=\linewidth]{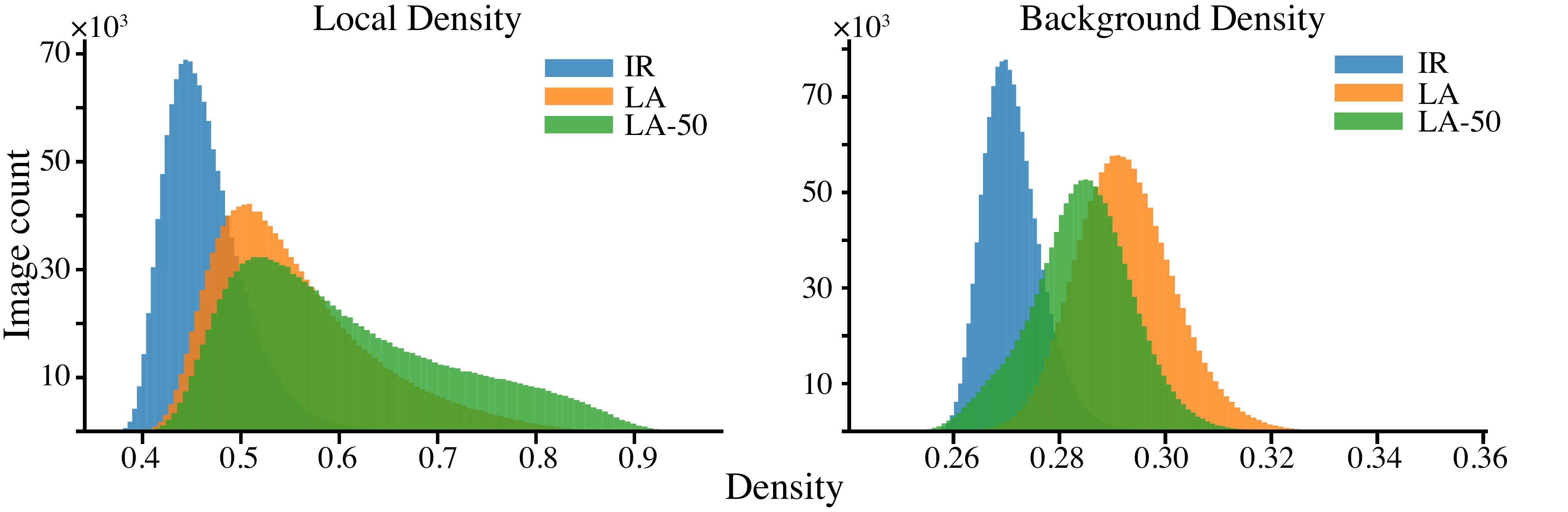}
\end{center}
\caption{
Distributions across all ImageNet training images of local and background densities for feature embeddings. We compare features from ResNet-18 (orange bars) and Resnet-50 (green bars) architectures as trained by the LA method, as well as that of a ResNet-18 architecture trained by the Instance Recognition (IR) method (blue bars).
The local and background densities at each embedded vector are estimated by averaging dot products between that vector and, respectively, its top 30 or its 1000th-4096th, nearest neighbors in $\MemBank$.
See supplementary material for more detail.
}
\label{fig:density}
\end{figure}

\textbf{Density distribution in the embedding space.} 
The LA optimization objective seeks to minimize the distances between $\mathbf{v}_i$ and $\SemNei$ while maximizing those between $\mathbf{v}_i$ and $\ConNei$, intuitively leading to an embedding that is locally dense at some positions but generally sparse across the space. 
Indeed, Figure~\ref{fig:density} shows that the local density of the LA embedding is much higher than that created by the IR method, while the background density is only slightly higher (note differing $x$-axis scales in the figure).
Moreover, insofar as deeper networks achieve lower minimums of the LA objective, we expect that their embeddings will exhibit higher local density and lower background density as compared to shallower networks.
By comparing the density distributions of the ResNet-18 embedding to that of ResNet-50, Figure~\ref{fig:density} shows that this expectation is confirmed.
These results help better characterize the LA optimization procedure.

\begin{figure}
\begin{center}
\includegraphics[width=\linewidth]{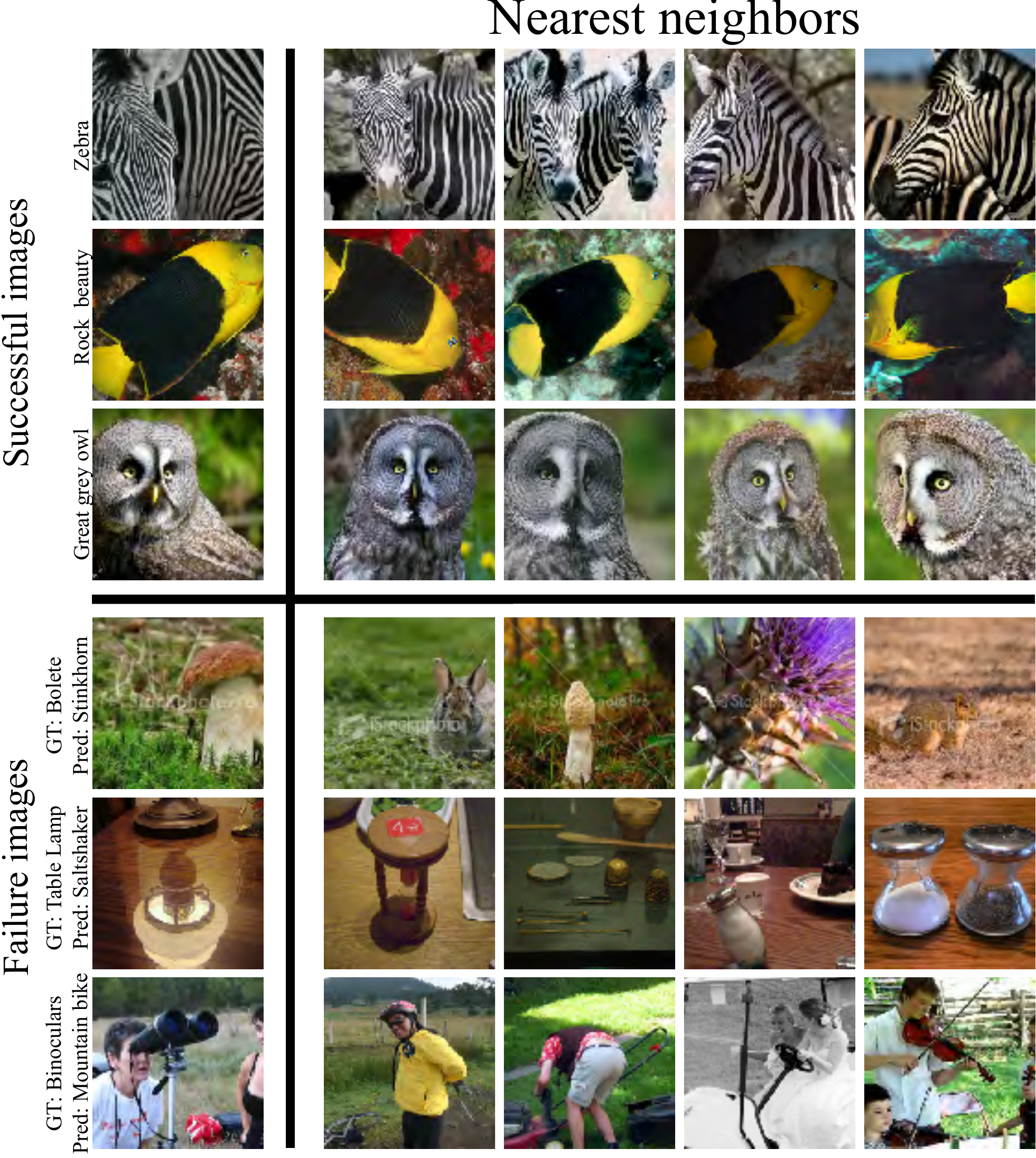}
\end{center}
\caption{
For each of several validation images in the left-most column, nearest neighbors in LA-trained RestNet-50 embedding, with similarity decreasing from left to right.
The three top columns are successfully-classified cases, with high KNN-classifier confidence, while the lower three are failure cases, with low KNN-classifier confidence.
}
\label{fig:visual}
\end{figure}

\begin{figure}
\begin{center}
\includegraphics[width=\linewidth]{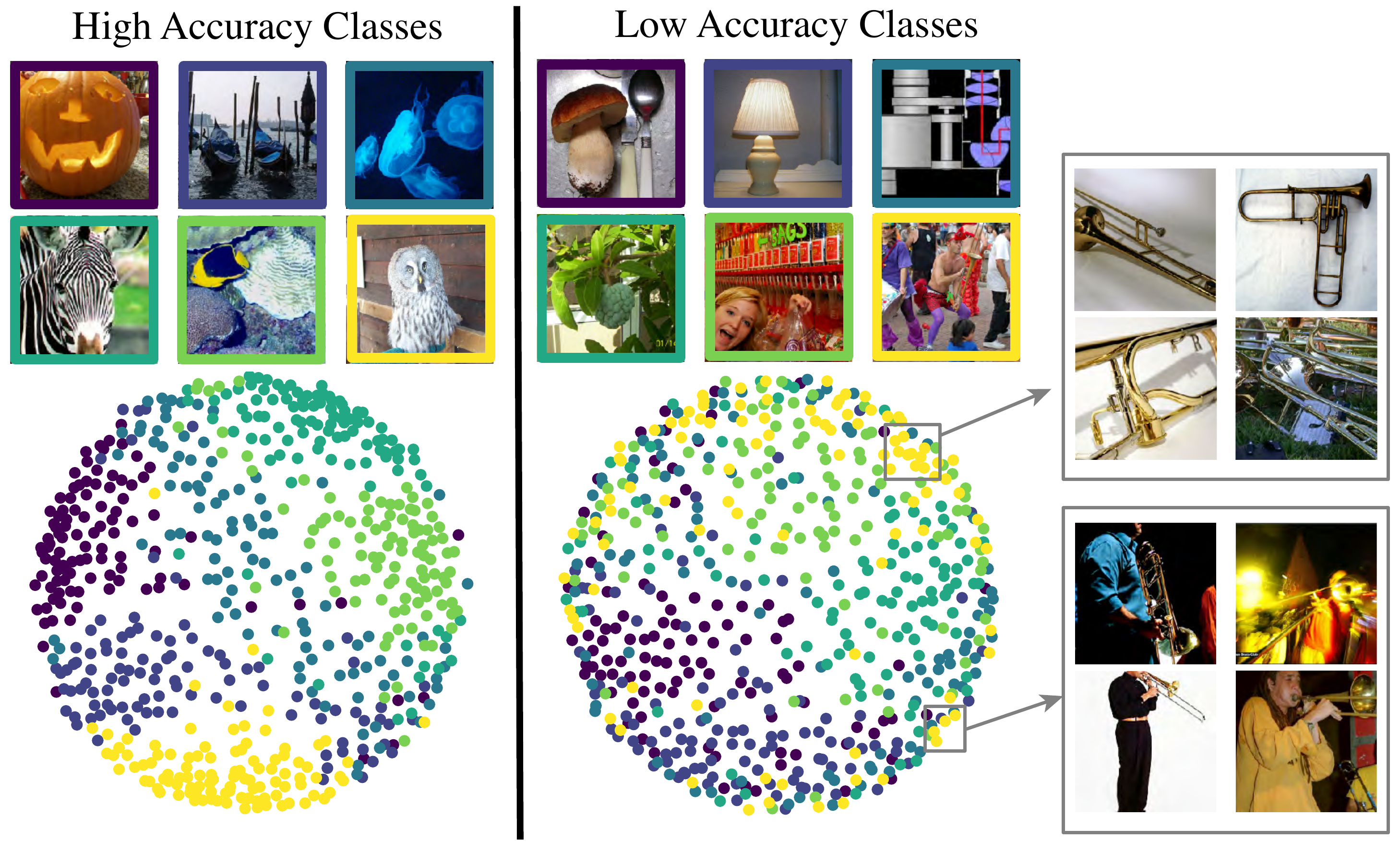}
\end{center}
\caption{
Multi-dimensional scaling (MDS) embedding results for network outputs of classes with high validation accuracy (left panel) and classes with low validation accuracy (right panel).
For each class, we randomly choose 100 images of that class from the training set and apply the MDS algorithm to the resulting 600 images.
Dots represent individual images in each color-coded category.
Gray boxes show examples of images from a single class ("trombone") that have been embedded in two distinct subclusters.
}
\label{fig:embed}
\end{figure}

\textbf{Success and failure examples.}
To help qualitatively illustrate the successes and failures of the LA objective, Figure~\ref{fig:visual} shows nearest neighbors in the training set for several validation images, both correctly and incorrectly classified according to the nearest-neighbor classifier.
Unsurprisingly, the successful examples show that the LA-trained model robustly groups images belonging to the same category regardless of backgrounds and view points.
Interestingly, however, the network shows substantial ability to recognize high-level visual context.
This is even more obvious for the failure cases, where it can be seen that the network coherently groups images according to salient characteristics.
In fact, most failure cases produced by the LA model appear to be due to the inherently ill-posed nature of the ImageNet category labelling, in which the category label is only one of several potentially valid object types present in the image, and which no unsupervised method could unambiguously resolve.
To further illustrate this point, we use the multi-dimensional scaling (MDS) algorithm~\cite{borg2003modern} to visualize part of the embedding space (see Fig.~\ref{fig:embed}). 
In particular, the LA successfully clusters images with trombones regardless of background, number of trombones, or viewpoint, while it (perhaps inevitably) distinguishes those images from images of humans playing trombones.

\subsection{Ablations}\label{sec:hyper}
In this subsection, we empirically justify the design of the LA procedure by ablating or modifying several key features of the procedure.
We also provide analyses suggesting intuitive reasons underlying the meaning and influence of parameters on final performance. Please refer to the supplementary material for further analyses.

\textbf{Dynamic Locality for Background Neighbors.}
\begin{table}
\begin{center}
\begin{tabular}{c|c|c|c}
\hline
Choice of $\ConNei$ & $\{1, 2, ..., N\}$ & Cluster-based & $\mathcal{N}_{4096}$  \\
\hline\hline
NN performance & 30.2 & 33.2 & \textbf{35.7} \\
\hline
\end{tabular}
\end{center}
\caption{
Nearest neighbor validation performances of ResNet-18 trained with different choices of $\ConNei$.
We use $H=3$ and $m=1000$ for cluster-based $\ConNei$ to make the number of neighbors in $\ConNei$ comparable to 4096.
In all experiments, we use cluster-based $\SemNei$ with $H=1$ and $m=10000$.
}
\label{tab:back_nei}
\end{table}
We chose a nearest-neighbor based procedure for identifying $\ConNei$ to embody the idea of dynamically rescaling the local background against which closeness is judged.
We tested two ablations of our procedure that isolate the relevance of this choice, including (i) simply using all inputs for background, or (ii) using a fixed clustering-based identification procedure. (See supplement for details on how these were defined.) 
Experiments show that the local dynamic nearest-neighbor procedure is substantially more performant than either ablation (see Table \ref{tab:back_nei}). The desirability of a local rather than global background measurement is consistent with the observation that the density of features varies widely across the embedding space (see Figure~\ref{fig:density}).
That the dynamic nature of the computation of the background is useful is illustrated by the comparison of results from computing neighbors in an online fashion from $\mathbf{v}_i$, relative to the cluster-based procedure depending only on $\MemBank$.

\textbf{Robust Clustering for Close Neighbors.}
\begin{table}
\begin{center}
\begin{tabular}{c|c|c|c|c|c|c}
\hline
$\SemNei$ & $\{i\}$ & $\mathcal{N}_{k'}$ & $(1, 10\mathrm{k})$ & $(3, 10\mathrm{k})$ & $(10, 10\mathrm{k})$ & $(10, 30\mathrm{k})$  \\
\hline\hline
NN & 33.9 & 0.1 & 35.7 & 36.2 & 36.1 & \textbf{37.9} \\
\hline
\end{tabular}
\end{center}
\caption{
Nearest neighbor validation performances of ResNet-18 trained with different choices of $\SemNei$.
All experiments use $\mathcal{N}_{4096}$ as $\ConNei$.
$\{i\}$ means $\SemNei$ only includes $\mathbf{v}_i$ itself.
$(1, 10\mathrm{k})$ means clustering-based $\SemNei$ with $H=1$ and $m=10000$.
Other pairs have similar meanings.
See the supplementary material for details.
}
\label{tab:close_nei}
\end{table}
We also sought to understand the importance of the specific clustering procedure for defining close neighbors $\SemNei$.
One alternative to using cluster-based identification would be to instead identify ``especially close'' neighbors as those within a neighborhood $\mathcal{N}_{k'}$, for some $k' \ll k$.  Using this in the definition of $\SemNei$ is equivalent to optimizing the embedding to bring especially close neighbors closer together, while somewhat further away neighbors are moved apart.
While this approach would have been a conceptually simpler way to define local aggregation than the cluster-based definition of close neighbors, it turns out to be substantially less effective in producing a useful representation (see Table \ref{tab:close_nei}).

Given the need for cluster-based identification, a variety of alternative approaches to $k$-means are theoretically possible, including DBSCAN~\cite{ester1996density}, Affinity Propagation~\cite{frey2007clustering}, spectral methods~\cite{shi2000normalized,stella2003multiclass}, and gaussian mixtures~\cite{reynolds2015gaussian,reynolds2000speaker}. 
However, our present context is strongly constrained by the requirement that the clustering algorithm scale well to large datasets, effectively limiting the options to $k$-means and DBSCAN.
Unfortunately, DBSCAN is known to perform poorly in settings with high ambient dimensions or highly variable density distributions~\cite{ester1996density}, both of which are characteristics of the embedding space we work with here (see Figure~\ref{fig:density}).
Indeed, we find that replacing $k$-means with DBSCAN leads to trivial representations, across a wide variety of parameter settings (see supplement for details). 

The robust clustering procedure described in Section \ref{sec:nei_iden} has several hyperparameters, including number of clusters $m$ and number of clusterings $H$.  
To intuitively understand their effect, we performed a set of network characterization experiments (see supplement for details).  
These experiments indicated that two basic factors were of importance in creating clusterings that lead to good representations: the \emph{skewness} of the cluster of close neighbors around its intended target, as measured by the distance from the cluster center to the embedded vector $\mathbf{v}_i$, and the \emph{size} of the cluster, as measured by its cardinality as a set.
We found that (i) clusterings of close neighbors with lower skewness were robustly associated with better performance, indicating that skewness should be minimized whenever possible; and (ii) there was an optimal size for the set of close neighbors that scaled with the representation capacity (i.e. depth) of the underlying network.  
Both of these facts are consistent with a picture in which the ideal embedding is one in which each category is equally likely to occur and in which each example of each category is equally ``representative'' -- e.g. in which clusters of points corresponding to natural categories occupy isotropic spheres of equal size.
Networks of smaller capacity that cannot completely achieve the optimal distribution will (poorly) approximate the optimal embedding by fracturing their embeddings of single categories into subsets that maintain isotropy by reducing the relative size of clusters, each containing only part of the true category. 
These considerations help explain the optimal settings for parameters $H$ and $m$: higher $H$ (i.e. more clusterings) will tend to produce more isotropic clusters, as outliers due to randomness are averaged out.  
However, increasing $H$ beyond a point set by the capacity of the network will lead to clusters of too large a size for the network to handle (see supplement Figure 1, from A to B, or from B to C).
This negative influence can be shown in Table~\ref{tab:close_nei} by the slight performance drop from $(3, 10\mathrm{k})$ to $(10, 10\mathrm{k})$.
Increasing $m$ (e.g. the number of clusters) can then compensate by decreasing the neighborhood size without increasing cluster anisotropy (see supplement Figure 1, from C to D).
This conpensation can be shown in Table~\ref{tab:close_nei} by the performance increase from $(10, 10\mathrm{k})$ to $(10, 30\mathrm{k})$.
More experiments detailing these conclusions are shown in the supplementary material.

\section{Discussion}
In this work, we have introduced a local aggregation (LA) objective for learning feature embeddings that seeks to discover a balance between bringing similar inputs together and allowing dissimilar inputs to move apart, embodying a principled combination of several key ideas from recent advances in unsupervised learning. We have shown that when applied to DCNNs, the LA objective creates representations that are useful for transfer learning to a variety of challenging visual tasks.  We also analyze aspects of our procedure, giving an intuition for how it works.

In future work we hope to improve the LA objective along a variety of directions, including incorporating non-local manifold learning-based priors for detecting similarity, improving identification of dissimilarity via measures of representational change over multiple steps of learning, and extending to the case of non-deterministic embedding functions. We also seek to apply the LA objective beyond the image processing domain, including to video and audio signals.  Finally, we hope to compare the LA procedure to biological vision systems, both in terms of the feature representations learned and the dynamics of learning during visual development.

\clearpage
{\small
\bibliographystyle{ieee}
\bibliography{egbib}

\begin{thebibliography}{10}\itemsep=-1pt

\bibitem{atkinson2002developing}
J.~Atkinson.
\newblock The developing visual brain.
\newblock 2002.

\bibitem{barlow1989unsupervised}
H.~B. Barlow.
\newblock Unsupervised learning.
\newblock {\em Neural computation}, 1(3):295--311, 1989.

\bibitem{borg2003modern}
I.~Borg and P.~Groenen.
\newblock Modern multidimensional scaling: Theory and applications.
\newblock {\em Journal of Educational Measurement}, 40(3):277--280, 2003.

\bibitem{bourne2005hierarchical}
J.~A. Bourne and M.~G. Rosa.
\newblock Hierarchical development of the primate visual cortex, as revealed by
  neurofilament immunoreactivity: early maturation of the middle temporal area
  (mt).
\newblock {\em Cerebral cortex}, 16(3):405--414, 2005.

\bibitem{braddick2011development}
O.~Braddick and J.~Atkinson.
\newblock Development of human visual function.
\newblock {\em Vision research}, 51(13):1588--1609, 2011.

\bibitem{caron2018deep}
M.~Caron, P.~Bojanowski, A.~Joulin, and M.~Douze.
\newblock Deep clustering for unsupervised learning of visual features.
\newblock In {\em Proceedings of the European Conference on Computer Vision
  (ECCV)}, pages 132--149, 2018.

\bibitem{carreira2017quo}
J.~Carreira and A.~Zisserman.
\newblock Quo vadis, action recognition? a new model and the kinetics dataset.
\newblock In {\em proceedings of the IEEE Conference on Computer Vision and
  Pattern Recognition}, pages 6299--6308, 2017.

\bibitem{cohen1979concept}
L.~B. Cohen and M.~S. Strauss.
\newblock Concept acquisition in the human infant.
\newblock {\em Child development}, pages 419--424, 1979.

\bibitem{conneau2016very}
A.~Conneau, H.~Schwenk, L.~Barrault, and Y.~Lecun.
\newblock Very deep convolutional networks for natural language processing.
\newblock {\em arXiv preprint arXiv:1606.01781}, 2, 2016.

\bibitem{deng2009imagenet}
J.~Deng, W.~Dong, R.~Socher, L.-J. Li, K.~Li, and L.~Fei-Fei.
\newblock Imagenet: A large-scale hierarchical image database.
\newblock 2009.

\bibitem{deng2013new}
L.~Deng, G.~Hinton, and B.~Kingsbury.
\newblock New types of deep neural network learning for speech recognition and
  related applications: An overview.
\newblock In {\em 2013 IEEE International Conference on Acoustics, Speech and
  Signal Processing}, pages 8599--8603. IEEE, 2013.

\bibitem{denton2015deep}
E.~L. Denton, S.~Chintala, R.~Fergus, et~al.
\newblock Deep generative image models using a￼ laplacian pyramid of
  adversarial networks.
\newblock In {\em Advances in neural information processing systems}, pages
  1486--1494, 2015.

\bibitem{doersch2015unsupervised}
C.~Doersch, A.~Gupta, and A.~A. Efros.
\newblock Unsupervised visual representation learning by context prediction.
\newblock In {\em Proceedings of the IEEE International Conference on Computer
  Vision}, pages 1422--1430, 2015.

\bibitem{doersch2017multi}
C.~Doersch and A.~Zisserman.
\newblock Multi-task self-supervised visual learning.
\newblock In {\em Proceedings of the IEEE International Conference on Computer
  Vision}, pages 2051--2060, 2017.

\bibitem{donahue2016adversarial}
J.~Donahue, P.~Kr{\"a}henb{\"u}hl, and T.~Darrell.
\newblock Adversarial feature learning.
\newblock {\em arXiv preprint arXiv:1605.09782}, 2016.

\bibitem{ester1996density}
M.~Ester, H.-P. Kriegel, J.~Sander, X.~Xu, et~al.
\newblock A density-based algorithm for discovering clusters in large spatial
  databases with noise.
\newblock In {\em Kdd}, volume~96, pages 226--231, 1996.

\bibitem{everingham2010pascal}
M.~Everingham, L.~Van~Gool, C.~K. Williams, J.~Winn, and A.~Zisserman.
\newblock The pascal visual object classes (voc) challenge.
\newblock {\em International journal of computer vision}, 88(2):303--338, 2010.

\bibitem{frey2007clustering}
B.~J. Frey and D.~Dueck.
\newblock Clustering by passing messages between data points.
\newblock {\em science}, 315(5814):972--976, 2007.

\bibitem{girshick2015fast}
R.~Girshick.
\newblock Fast r-cnn.
\newblock In {\em Proceedings of the IEEE international conference on computer
  vision}, pages 1440--1448, 2015.

\bibitem{goodfellow2014generative}
I.~Goodfellow, J.~Pouget-Abadie, M.~Mirza, B.~Xu, D.~Warde-Farley, S.~Ozair,
  A.~Courville, and Y.~Bengio.
\newblock Generative adversarial nets.
\newblock In {\em Advances in neural information processing systems}, pages
  2672--2680, 2014.

\bibitem{hannun2014deep}
A.~Hannun, C.~Case, J.~Casper, B.~Catanzaro, G.~Diamos, E.~Elsen, R.~Prenger,
  S.~Satheesh, S.~Sengupta, A.~Coates, et~al.
\newblock Deep speech: Scaling up end-to-end speech recognition.
\newblock {\em arXiv preprint arXiv:1412.5567}, 2014.

\bibitem{harwerth1986multiple}
R.~S. Harwerth, E.~L. Smith, G.~C. Duncan, M.~Crawford, and G.~K. Von~Noorden.
\newblock Multiple sensitive periods in the development of the primate visual
  system.
\newblock {\em Science}, 232(4747):235--238, 1986.

\bibitem{he2017mask}
K.~He, G.~Gkioxari, P.~Doll{\'a}r, and R.~Girshick.
\newblock Mask r-cnn.
\newblock In {\em Proceedings of the IEEE international conference on computer
  vision}, pages 2961--2969, 2017.

\bibitem{he2016deep}
K.~He, X.~Zhang, S.~Ren, and J.~Sun.
\newblock Deep residual learning for image recognition.
\newblock In {\em Proceedings of the IEEE conference on computer vision and
  pattern recognition}, pages 770--778, 2016.

\bibitem{hebb1961organization}
D.~O. Hebb.
\newblock {\em The organization of behavior}.
\newblock na, 1961.

\bibitem{hinton2012deep}
G.~Hinton, L.~Deng, D.~Yu, G.~Dahl, A.-r. Mohamed, N.~Jaitly, A.~Senior,
  V.~Vanhoucke, P.~Nguyen, B.~Kingsbury, et~al.
\newblock Deep neural networks for acoustic modeling in speech recognition.
\newblock {\em IEEE Signal processing magazine}, 29, 2012.

\bibitem{hinton2006fast}
G.~E. Hinton, S.~Osindero, and Y.-W. Teh.
\newblock A fast learning algorithm for deep belief nets.
\newblock {\em Neural computation}, 18(7):1527--1554, 2006.

\bibitem{hinton2006reducing}
G.~E. Hinton and R.~R. Salakhutdinov.
\newblock Reducing the dimensionality of data with neural networks.
\newblock {\em science}, 313(5786):504--507, 2006.

\bibitem{hirschberg2015advances}
J.~Hirschberg and C.~D. Manning.
\newblock Advances in natural language processing.
\newblock {\em Science}, 349(6245):261--266, 2015.

\bibitem{hopfield1982neural}
J.~J. Hopfield.
\newblock Neural networks and physical systems with emergent collective
  computational abilities.
\newblock {\em Proceedings of the national academy of sciences},
  79(8):2554--2558, 1982.

\bibitem{husaim1981infant}
J.~S. Husaim and L.~B. Cohen.
\newblock Infant learning of ill-defined categories.
\newblock {\em Merrill-Palmer Quarterly of Behavior and Development}, pages
  443--456, 1981.

\bibitem{ioffe2015batch}
S.~Ioffe and C.~Szegedy.
\newblock Batch normalization: Accelerating deep network training by reducing
  internal covariate shift.
\newblock {\em arXiv preprint arXiv:1502.03167}, 2015.

\bibitem{jetchev2016texture}
N.~Jetchev, U.~Bergmann, and R.~Vollgraf.
\newblock Texture synthesis with spatial generative adversarial networks.
\newblock {\em arXiv preprint arXiv:1611.08207}, 2016.

\bibitem{JDH17}
J.~Johnson, M.~Douze, and H.~J{\'e}gou.
\newblock Billion-scale similarity search with gpus.
\newblock {\em arXiv preprint arXiv:1702.08734}, 2017.

\bibitem{kingma2013auto}
D.~P. Kingma and M.~Welling.
\newblock Auto-encoding variational bayes.
\newblock {\em arXiv preprint arXiv:1312.6114}, 2013.

\bibitem{krahenbuhl2015data}
P.~Kr{\"a}henb{\"u}hl, C.~Doersch, J.~Donahue, and T.~Darrell.
\newblock Data-dependent initializations of convolutional neural networks.
\newblock {\em arXiv preprint arXiv:1511.06856}, 2015.

\bibitem{krizhevsky2012imagenet}
A.~Krizhevsky, I.~Sutskever, and G.~E. Hinton.
\newblock Imagenet classification with deep convolutional neural networks.
\newblock In {\em Advances in neural information processing systems}, pages
  1097--1105, 2012.

\bibitem{kumar2016ask}
A.~Kumar, O.~Irsoy, P.~Ondruska, M.~Iyyer, J.~Bradbury, I.~Gulrajani, V.~Zhong,
  R.~Paulus, and R.~Socher.
\newblock Ask me anything: Dynamic memory networks for natural language
  processing.
\newblock In {\em International Conference on Machine Learning}, pages
  1378--1387, 2016.

\bibitem{le2011building}
Q.~V. Le, M.~Ranzato, R.~Monga, M.~Devin, K.~Chen, G.~S. Corrado, J.~Dean, and
  A.~Y. Ng.
\newblock Building high-level features using large scale unsupervised learning.
\newblock {\em arXiv preprint arXiv:1112.6209}, 2011.

\bibitem{ledig2017photo}
C.~Ledig, L.~Theis, F.~Husz{\'a}r, J.~Caballero, A.~Cunningham, A.~Acosta,
  A.~Aitken, A.~Tejani, J.~Totz, Z.~Wang, et~al.
\newblock Photo-realistic single image super-resolution using a generative
  adversarial network.
\newblock In {\em Proceedings of the IEEE conference on computer vision and
  pattern recognition}, pages 4681--4690, 2017.

\bibitem{lee2009convolutional}
H.~Lee, R.~Grosse, R.~Ranganath, and A.~Y. Ng.
\newblock Convolutional deep belief networks for scalable unsupervised learning
  of hierarchical representations.
\newblock In {\em Proceedings of the 26th annual international conference on
  machine learning}, pages 609--616. ACM, 2009.

\bibitem{lewis2005multiple}
T.~L. Lewis and D.~Maurer.
\newblock Multiple sensitive periods in human visual development: evidence from
  visually deprived children.
\newblock {\em Developmental Psychobiology: The Journal of the International
  Society for Developmental Psychobiology}, 46(3):163--183, 2005.

\bibitem{li2016precomputed}
C.~Li and M.~Wand.
\newblock Precomputed real-time texture synthesis with markovian generative
  adversarial networks.
\newblock In {\em European Conference on Computer Vision}, pages 702--716.
  Springer, 2016.

\bibitem{lloyd1982least}
S.~Lloyd.
\newblock Least squares quantization in pcm.
\newblock {\em IEEE transactions on information theory}, 28(2):129--137, 1982.

\bibitem{malisiewicz2011ensemble}
T.~Malisiewicz, A.~Gupta, and A.~A. Efros.
\newblock Ensemble of exemplar-svms for object detection and beyond.
\newblock 2011.

\bibitem{mareschal2001categorization}
D.~Mareschal and P.~C. Quinn.
\newblock Categorization in infancy.
\newblock {\em Trends in cognitive sciences}, 5(10):443--450, 2001.

\bibitem{noda2015audio}
K.~Noda, Y.~Yamaguchi, K.~Nakadai, H.~G. Okuno, and T.~Ogata.
\newblock Audio-visual speech recognition using deep learning.
\newblock {\em Applied Intelligence}, 42(4):722--737, 2015.

\bibitem{noroozi2016unsupervised}
M.~Noroozi and P.~Favaro.
\newblock Unsupervised learning of visual representations by solving jigsaw
  puzzles.
\newblock In {\em European Conference on Computer Vision}, pages 69--84.
  Springer, 2016.

\bibitem{noroozi2017representation}
M.~Noroozi, H.~Pirsiavash, and P.~Favaro.
\newblock Representation learning by learning to count.
\newblock In {\em Proceedings of the IEEE International Conference on Computer
  Vision}, pages 5898--5906, 2017.

\bibitem{oord2018representation}
A.~v.~d. Oord, Y.~Li, and O.~Vinyals.
\newblock Representation learning with contrastive predictive coding.
\newblock {\em arXiv preprint arXiv:1807.03748}, 2018.

\bibitem{pathak2017learning}
D.~Pathak, R.~Girshick, P.~Doll{\'a}r, T.~Darrell, and B.~Hariharan.
\newblock Learning features by watching objects move.
\newblock In {\em Proceedings of the IEEE Conference on Computer Vision and
  Pattern Recognition}, pages 2701--2710, 2017.

\bibitem{pathak2016context}
D.~Pathak, P.~Krahenbuhl, J.~Donahue, T.~Darrell, and A.~A. Efros.
\newblock Context encoders: Feature learning by inpainting.
\newblock In {\em Proceedings of the IEEE conference on computer vision and
  pattern recognition}, pages 2536--2544, 2016.

\bibitem{radford2015unsupervised}
A.~Radford, L.~Metz, and S.~Chintala.
\newblock Unsupervised representation learning with deep convolutional
  generative adversarial networks.
\newblock {\em arXiv preprint arXiv:1511.06434}, 2015.

\bibitem{rakison2003early}
D.~H. Rakison and L.~M. Oakes.
\newblock {\em Early category and concept development: Making sense of the
  blooming, buzzing confusion}.
\newblock Oxford University Press, 2003.

\bibitem{renNIPS15fasterrcnn}
S.~Ren, K.~He, R.~Girshick, and J.~Sun.
\newblock Faster {R-CNN}: Towards real-time object detection with region
  proposal networks.
\newblock In {\em Advances in Neural Information Processing Systems ({NIPS})},
  2015.

\bibitem{reynolds2015gaussian}
D.~Reynolds.
\newblock Gaussian mixture models.
\newblock {\em Encyclopedia of biometrics}, pages 827--832, 2015.

\bibitem{reynolds2000speaker}
D.~A. Reynolds, T.~F. Quatieri, and R.~B. Dunn.
\newblock Speaker verification using adapted gaussian mixture models.
\newblock {\em Digital signal processing}, 10(1-3):19--41, 2000.

\bibitem{sanger1989optimal}
T.~D. Sanger.
\newblock Optimal unsupervised learning in a single-layer linear feedforward
  neural network.
\newblock {\em Neural networks}, 2(6):459--473, 1989.

\bibitem{shi2000normalized}
J.~Shi and J.~Malik.
\newblock Normalized cuts and image segmentation.
\newblock {\em Departmental Papers (CIS)}, page 107, 2000.

\bibitem{simonyan2014very}
K.~Simonyan and A.~Zisserman.
\newblock Very deep convolutional networks for large-scale image recognition.
\newblock {\em arXiv preprint arXiv:1409.1556}, 2014.

\bibitem{sporns2004organization}
O.~Sporns, D.~R. Chialvo, M.~Kaiser, and C.~C. Hilgetag.
\newblock Organization, development and function of complex brain networks.
\newblock {\em Trends in cognitive sciences}, 8(9):418--425, 2004.

\bibitem{stella2003multiclass}
X.~Y. Stella and J.~Shi.
\newblock Multiclass spectral clustering.
\newblock In {\em null}, page 313. IEEE, 2003.

\bibitem{wang2015unsupervised}
X.~Wang and A.~Gupta.
\newblock Unsupervised learning of visual representations using videos.
\newblock In {\em Proceedings of the IEEE International Conference on Computer
  Vision}, pages 2794--2802, 2015.

\bibitem{wang2017transitive}
X.~Wang, K.~He, and A.~Gupta.
\newblock Transitive invariance for self-supervised visual representation
  learning.
\newblock In {\em Proceedings of the IEEE International Conference on Computer
  Vision}, pages 1329--1338, 2017.

\bibitem{wattam2010reorganization}
J.~Wattam-Bell, D.~Birtles, P.~Nystr{\"o}m, C.~Von~Hofsten, K.~Rosander,
  S.~Anker, J.~Atkinson, and O.~Braddick.
\newblock Reorganization of global form and motion processing during human
  visual development.
\newblock {\em Current Biology}, 20(5):411--415, 2010.

\bibitem{wu2018improving}
Z.~Wu, A.~A. Efros, and S.~X. Yu.
\newblock Improving generalization via scalable neighborhood component
  analysis.
\newblock In {\em Proceedings of the European Conference on Computer Vision
  (ECCV)}, pages 685--701, 2018.

\bibitem{wu2018unsupervised}
Z.~Wu, Y.~Xiong, S.~X. Yu, and D.~Lin.
\newblock Unsupervised feature learning via non-parametric instance
  discrimination.
\newblock In {\em Proceedings of the IEEE Conference on Computer Vision and
  Pattern Recognition}, pages 3733--3742, 2018.

\bibitem{young2018recent}
T.~Young, D.~Hazarika, S.~Poria, and E.~Cambria.
\newblock Recent trends in deep learning based natural language processing.
\newblock {\em ieee Computational intelligenCe magazine}, 13(3):55--75, 2018.

\bibitem{zhang2017stackgan}
H.~Zhang, T.~Xu, H.~Li, S.~Zhang, X.~Wang, X.~Huang, and D.~N. Metaxas.
\newblock Stackgan: Text to photo-realistic image synthesis with stacked
  generative adversarial networks.
\newblock In {\em Proceedings of the IEEE International Conference on Computer
  Vision}, pages 5907--5915, 2017.

\bibitem{zhang2016colorful}
R.~Zhang, P.~Isola, and A.~A. Efros.
\newblock Colorful image colorization.
\newblock In {\em European conference on computer vision}, pages 649--666.
  Springer, 2016.

\bibitem{zhang2017split}
R.~Zhang, P.~Isola, and A.~A. Efros.
\newblock Split-brain autoencoders: Unsupervised learning by cross-channel
  prediction.
\newblock In {\em Proceedings of the IEEE Conference on Computer Vision and
  Pattern Recognition}, pages 1058--1067, 2017.

\bibitem{zhou2014learning}
B.~Zhou, A.~Lapedriza, J.~Xiao, A.~Torralba, and A.~Oliva.
\newblock Learning deep features for scene recognition using places database.
\newblock In {\em Advances in neural information processing systems}, pages
  487--495, 2014.

\end{thebibliography}
}

\clearpage
\appendix
\setcounter{figure}{0}
\setcounter{table}{0}

{\Large \textbf{Supplementary Material}}

\section{Clustering Combination}
\begin{figure}[ht]
\begin{center}
\includegraphics[width=\linewidth]{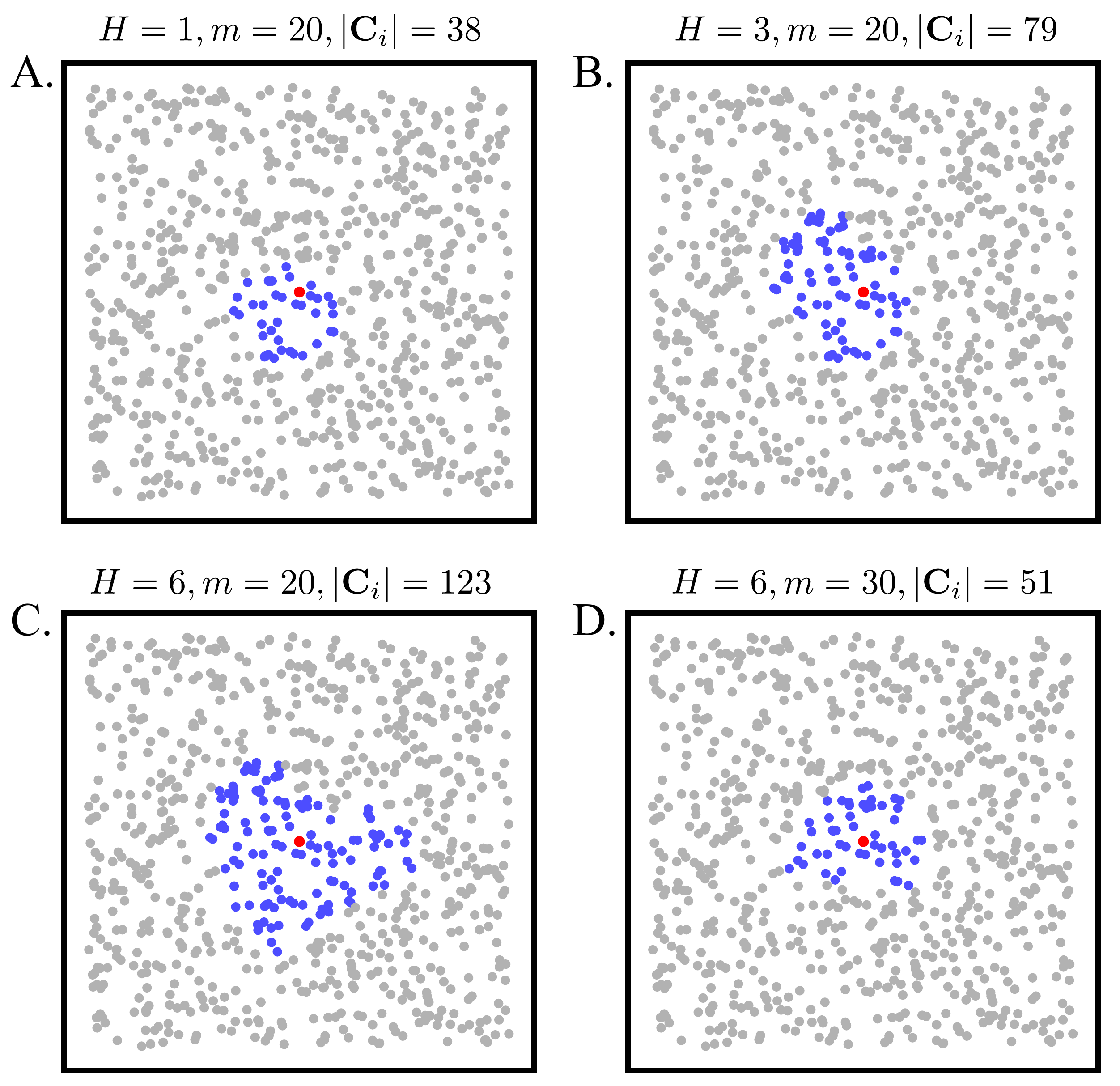}
\end{center}
\caption{
Illustration of the effect of combining across multiple clusterings to achieve robustness.
The target embedded vector $\mathbf{v}_i$ is represented by the red dot, while blue dots represent close neighbors $\SemNei$ under the specified hyperparameter settings.
}
\label{fig:clster}
\end{figure}

In this section we provide an illustration figure for the effects of combining multiple clusterings in Figure~\ref{fig:clster}, which is also mentioned in Section 5.2. 
Additionally, we show the nearest neighbor validation performances in Table~\ref{tab:combine} to support our hyper-parameter choices for $H, m$ in different architectures.

\begin{table}[ht]
\begin{center}
\begin{tabular}{c|cccc}
\hline
\diagbox[width=7em]{Setting}{Network} & A & V & R-18 & R-50 \\
\hline\hline
$(1, 1\mathrm{k})$ & -- & -- & 35.2 & -- \\
\hline
$(1, 10\mathrm{k})$ & 30.6 & 38.9 & 35.7 & 40.2 \\
\hline
$(1, 20\mathrm{k})$ & -- & -- & 35.0 & -- \\
\hline
$(3, 10\mathrm{k})$ & \textbf{31.1} & -- & 36.2 & --\\
\hline
$(6, 10\mathrm{k})$ & 30.4 & \textbf{39.7} & 37.3 & 42.4 \\
\hline
$(10, 10\mathrm{k})$ & -- & -- & 36.1 & 42.3 \\
\hline
$(10, 30\mathrm{k})$ & -- & -- & \textbf{37.9} & \textbf{43.4} \\
\hline
\end{tabular}
\end{center}
\caption{
Nearest neighbor validation performances of different architectures trained with different choices of $\SemNei$.
``A'' means ``AlexNet''. ``V'' means VGG16. ``R'' means ``ResNet''.
Similarly to Table 5 in the main text, $(1, 10\mathrm{k})$ means clustering-based $\SemNei$ with $H=1$ and $m=10000$.
}
\label{tab:combine}
\end{table}

\section{Results Details}
\subsection{Transfer Learning Details}\label{sec:trans}
Besides the settings listed in the main paper, there are additional settings for data augmentation during our transfer learning training to ImageNet and Places 205 datasets. 
In general, we use random crop and random horizontal flip as data augmentation techniques during transfer learning for all architectures on both ImageNet and Places 205 datasets, where the specific random crop implementation varies across networks and datasets.
For AlexNet on ImageNet and all architectures on Places 205, we use the AlexNet style random crop~\cite{hinton2012deep}, which is first resizing the image so that its smallest side is 256 and then randomly cropping a $224 \times 224$ patch.
For VGG16, ResNet-18, and ResNet-50 on ImageNet, we use the ResNet style random crop~\cite{he2016deep}, which is first randomly choosing a patch whose aspect ratio and area suffice two conditions and then resizing that path to $224 \times 224$. 
The two sufficed conditions are: its area is at least $20\%$ of the overall area and at most $100\%$ of the overall area; its aspect ratio ranges from $3 / 4$ to $4 / 3$.
We use the same data augmentation techniques for the same architecture trained with different methods.

\subsection{DeepCluster Results Details}
The DeepCluster~\cite{caron2018deep} VGG16, ResNet-18, and ResNet-50 results are produced by us, where the DC-VGG16 network is provided by the authors and the DC-ResNet-18 and DC-ResNet-50 networks are trained by us using the provided source codes. 

More specifically, for ResNet-18, two implementations of DC-ResNet-18 network are trained. Both of them modifies the standard ResNet-18 architecture by removing the final pooling and final fully connected layer and then adding additional fully connected layers, where the last layer has 10000 units.
One implementation (DC-ResNet-18-A) only has that 10000-unit fully connected layer and the other implementation (DC-ResNet-18-B) has two more 4096-unit fully connected layers before that.
We find that DC-ResNet-18-B performs slightly better than DC-ResNet-18-A and thus report the performances of DC-ResNet-18-B in the main paper.

Similarly for ResNet-50, two implementations (DC-ResNet-50-A and DC-ResNet-50-B) are trained. 
However, we find it impossible to train DC-ResNet-50-B as the $k$-means clustering results always become trivial at the third epoch.
So the results reported in the paper are from DC-ResNet-50-A, which should only be slightly worse than DC-ResNet-50-B.

Other hyper-parameters for network training are mostly the same as used in the provided source codes.
Meanwhile, all hyper-parameters for transfer learning to ImageNet and Places 205 are also the same as provided, except the data augmentation techniques which are the same as described in Section~\ref{sec:trans}.

\subsection{Places KNN Results}
We run models on center crops of training images in Places 205~\cite{zhou2014learning} dataset to generate the memory bank $\MemBank$.
We then run the KNN validation similarly to the ImageNet~\cite{deng2009imagenet} KNN procedure, which is described in the main paper.
The results are shown in Table~\ref{tab:place_knn}.
\begin{table}[ht]
\begin{center}
\begin{tabular}{c|c}
\hline
Network & KNN \\
\hline\hline
IR with BN - A & 36.9 \\
\hline
LA - A & \textbf{37.5} \\
\hline\hline
IR with BN - V & 40.1 \\
\hline
LA - V & \textbf{41.9} \\
\hline\hline
IR - R18 & 38.6 \\ 
\hline
LA - R18 & \textbf{40.3} \\
\hline\hline
IR - R50 & 41.6 \\ 
\hline
LA - R50 & \textbf{42.4} \\
\hline
\end{tabular}
\end{center}
\caption{
KNN results for Places 205 dataset. 
``A'' means ``AlexNet''. ``V'' means VGG16. ``R'' means ``ResNet''.
}
\label{tab:place_knn}
\end{table}

\subsection{Faster RCNN Details}
Our Faster RCNN~\cite{renNIPS15fasterrcnn} implementations are based on \href{https://github.com/endernewton/tf-faster-rcnn}{tf-faster-rcnn}.
We use SGD with momentum of 0.9, batch size 256, and weight decay 0.0001.
Learning rate is initialized as 0.001 and dropped by a factor of 10 after 50000 steps.
We train the models for 70000 steps.
In particular, we set the number of total RoIs for training the region classifier to be 128 to reproduce the original Faster RCNN results, as indicated by~\cite{chen17implementation}.
For AlexNet, we fine-tune all layers.
For VGG16, we fix ``conv1'' and ``conv2'' while fine-tuning others.
For ResNet-50, we fix the first convolution layer and the first three blocks while fine-tuning others.
Other hyper-parameters are the same as the default settings in \href{https://github.com/endernewton/tf-faster-rcnn}{tf-faster-rcnn}.

\section{Other Hyperparameters}
There are several other adjustable hyper-parameters in LA training procedure, such as the updating frequency for the clustering results, the parameter $k$ in $\mathcal{N}_k$ for $\ConNei$, and whether doing clustering on $\MemBank$ or network outputs on center crops of $\mathbf{I}$. 
In this section, we show results of experiments illustrating the influences of these parameters in Table~\ref{tab:other_hyper}.

\begin{table}[ht]
\begin{center}
\begin{tabular}{c|c}
\hline
Setting & NN perf. \\
\hline\hline
Baseline & 35.7 \\
\hline
$k=2048$ & 35.4 \\
\hline\hline
$k=8192$ & 35.8 \\
\hline
center$\_$crop & 35.8 \\
\hline\hline
more$\_$freq & 35.7 \\ 
\hline
\end{tabular}
\end{center}
\caption{
Nearest neighbor validation performances for ResNet-18 trained with different settings. 
``Baseline'' uses $H=1, m=10000,$ and $k=4096$.
Other settings change one of the hyper-parameters while keeping the others the same.
``center$\_$crop'' represents the experiment with clustering result acquired on the center crops rather than $\MemBank$.
``more$\_$freq'' represents the experiment with clustering result updated every 1000 steps.
}
\label{tab:other_hyper}
\end{table}

\end{document}